\documentclass{article}
\usepackage[final]{neurips_2023}
\usepackage{hyperref}
\usepackage{bbding}

\usepackage{natbib}
\setcitestyle{numbers,square}

\usepackage[utf8]{inputenc} 
\usepackage[T1]{fontenc}    
\usepackage{hyperref}       
\usepackage{url}            
\usepackage{booktabs}       
\usepackage{amsfonts}       
\usepackage{nicefrac}       
\usepackage{microtype}      
\usepackage{xcolor}         
\usepackage{multirow}
\usepackage{graphicx}
\usepackage{CJKutf8}
\usepackage[many]{tcolorbox}
\usepackage{graphicx}
\usepackage{geometry}
\usepackage{tabularx}
\usepackage{booktabs}
\usepackage{makecell}
\usepackage{cleveref}

\definecolor{green}{RGB}{0,150,10}
\definecolor{blue}{RGB}{0,148,181}
\definecolor{orange}{RGB}{194,153,107}

\title{MinerU: An Open-Source Solution for Precise Document Content Extraction}

\author{
Bin Wang\footnotemark[1],\quad
Chao Xu\thanks{Project leader.},\quad 
Xiaomeng Zhao,\quad
Linke Ouyang,\quad
\\\textbf{Fan Wu},\quad
\textbf{Zhiyuan Zhao,}\quad
\textbf{Rui Xu,}\quad
\textbf{Kaiwen Liu,}\quad
\textbf{Yuan Qu,}\quad
\\\textbf{Fukai Shang,}\quad
\textbf{Bo Zhang,}\quad
\textbf{Liqun Wei,}\quad
\textbf{Zhihao Sui,}\quad
\textbf{Wei Li,}\quad
\\\textbf{Botian Shi,}\quad
\textbf{Yu Qiao,}\quad
\textbf{Dahua Lin,}\quad
\textbf{Conghui He}\thanks{Corresponding author: heconghui@pjlab.org.cn}
\\[1.5mm]
Shanghai Artificial Intelligence Laboratory}
    
\begin{document}

\maketitle

\begin{abstract}
Document content analysis has been a crucial research area in computer vision. Despite significant advancements in methods such as OCR, layout detection, and formula recognition, existing open-source solutions struggle to consistently deliver high-quality content extraction due to the diversity in document types and content. To address these challenges, we present MinerU, an open-source solution for high-precision document content extraction. MinerU leverages the sophisticated PDF-Extract-Kit models to extract content from diverse documents effectively and employs finely-tuned preprocessing and postprocessing rules to ensure the accuracy of the final results. Experimental results demonstrate that MinerU consistently achieves high performance across various document types, significantly enhancing the quality and consistency of content extraction. The MinerU open-source project is available at \url{https://github.com/opendatalab/MinerU}.

\end{abstract}

\section{Introduction}

The release of ChatGPT~\cite{chatgpt,openai2023gpt4} at the end of 2022 ignites a wave of interest in the research and application of large language models (LLMs)~\cite{li2023starcoder,roziere2023codellama,team2023internlm,ying2024internlm,cai2024internlm2,llama2,llama3,gpt3,gpt4,mistral}. Central to training high-quality LLMs is the acquisition and construction of high-quality data. As LLMs rapidly evolve, data from internet web pages is becoming insufficient to support further improvements in model training. Document data, which contains a wealth of knowledge, emerges as a crucial resource for enhancing LLMs. The introduction and development of Retrieval-Augmented Generation (RAG) ~\cite{lewis2020retrieval,ram2023context,asai2023self,edge2024local} in 2023 further intensify the demand for high-quality document extraction in both industry and academia.

Currently, there are four main technical approaches to document content extraction:

\textbf{OCR-based Text Extraction}. This approach uses OCR models to directly extract text from documents. While feasible for plain text documents, it introduces significant noise when documents contain images, tables, formulas, and other elements, rendering it unsuitable for high-quality data extraction.
    
\textbf{Library-based Text Parsing}. For non-scanned documents, open-source Python libraries such as PyMuPDF can directly read content without invoking OCR, offering faster and more accurate text results. However, this approach fails when documents contain formulas, tables, and other elements.

\textbf{Multi-Module Document Parsing}. This approach employs various document parsing models to process document images in multiple stages. Initially, layout detection algorithms identify different types of regions, such as images, image captions, tables, table captions, headings, and text. Subsequently, different recognizers are applied to these specific regions. For instance, OCR is used for text and headings, formula recognition models handle formulas, and table recognition models process tables. Although this method is theoretically capable of producing high-quality document results, existing open-source models often focus solely on academic papers and perform poorly on diverse document types, including textbooks, exam papers, research reports, and newspapers.

\textbf{End-to-End MLLM Document Parsing}. With the advancement of multimodal large language models (MLLMs), numerous models for document content extraction emerge, such as Donut~\cite{kim2022ocr}, Nougat~\cite{blecher2023nougat}, Kosmos-2.5~\cite{lv2023kosmos}, Vary~\cite{wei2023vary}, Vary-toy~\cite{wei2024small}, mPLUG-DocOwl-1.5~\cite{hu2024mplug}, mPLUG-DocOwl2~\cite{hu2024mplug-2.0}, Fox~\cite{liu2024focus}, and GOT~\cite{wei2024general}. These models benefit from continuously optimized encoders (e.g., SwinTransformer~\cite{liu2021swin}, ViTDet~\cite{li2022exploring}) and decoders (e.g., mBART~\cite{liu2020multilingual}, Qwen2-0.5B~\cite{yang2024qwen2}) as well as data engineering, gradually improving extraction performance. However, they still face challenges related to data diversity and high inference costs.

To better extract diverse documents while ensuring low inference costs and high inference quality, we propose MinerU, an all-in-one document extraction tool. MinerU's primary technical approach is based on the multi-module document parsing strategy. Unlike existing document parsing algorithms, MinerU leverages various open-source models from the PDF-Extract-Kit\footnote{\url{https://github.com/opendatalab/PDF-Extract-Kit}}, which are trained on diverse real-world documents to achieve high-quality results in tasks involving complex layouts and intricate formulas. After obtaining the positions and recognition content of different regions from the models, MinerU employs a tailored processing workflow to ensure the accuracy of the results.

Using MinerU for document extraction offers several advantages:

\begin{itemize}
    \item \textbf{Adaptability to Diverse Document Layouts}: Supports a wide range of document types, including but not limited to academic papers, textbooks, exam papers, and research reports.
    \item \textbf{Content Filtering}: Allows filtering of irrelevant regions such as headers, footers, footnotes, and side notes, enhancing document readability.
    \item \textbf{Accurate Segmentation}: Combines model-based and rule-based post-processing for paragraph recognition, enabling cross-column and cross-page paragraph merging.
    \item \textbf{Robust Page Element Recognition}: Accurately distinguishes between formulas, tables, images, text blocks, and their respective captions.
\end{itemize}

\begin{figure*}[t]
\centering
    \includegraphics[width=0.95\linewidth]{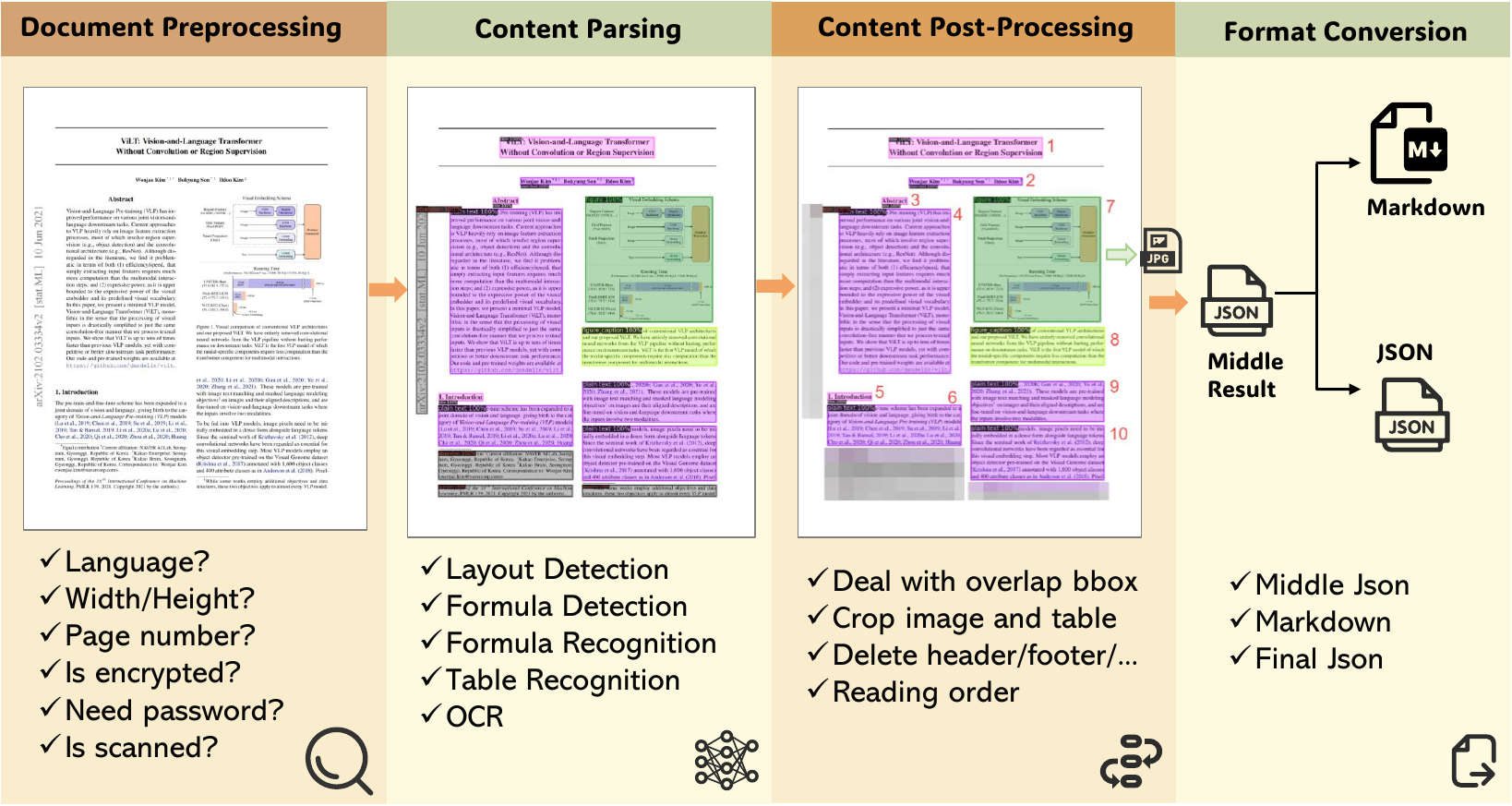}
    \caption{Overview of the MinerU framework processing workflow.}
\label{fig:fig1}
\vspace{-15pt}
\end{figure*}

\section{MinerU Framework}

As shown in \Cref{fig:fig1}, MinerU processes diverse user-input PDF documents and converts them into desired machine-readable formats (Markdown or JSON) through a series of steps. Specifically, the processing workflow of MinerU is divided into four stages:

\textbf{Document Preprocessing}. This stage uses PyMuPDF\footnote{\url{https://github.com/pymupdf/PyMuPDF}}  to read PDF files, filter out unprocessable files (e.g., encrypted files), and extract PDF metadata, including the document's parseability (categorized into parseable and scanned PDFs), language type, and page dimensions.

\textbf{Document Content Parsing}. This stage employs the PDF-Extract-Kit, a high-quality PDF document extraction algorithm library, to parse key document contents. It begins with layout analysis, including layout and formula detection. Different recognizers are then applied to various regions: OCR~\cite{smith2007overview,liu2023hidden} for text and titles, formula recognition~\cite{pix2tex2022,texify2023,wang2024cdm} for formulas, and table recognition~\cite{xia2024docgenome} for tables.

\textbf{Document Content Post-Processing}. Based on the outputs from the second stage, this stage removes invalid regions, stitches content according to regional positioning information, and ultimately obtains the positioning, content, and sorting information for different document regions.

\textbf{Format Conversion}. Based on the results of document post-processing, various formats required by users, such as Markdown, can be generated for subsequent use.

\subsection{Document Preprocessing}

PDF document preprocessing has two main objectives: first, to filter out unprocessable PDFs, such as non-PDF files, encrypted documents, and password-protected documents. Second, to obtain PDF metadata for subsequent use. The acquisition of PDF metadata includes the following aspects:

\begin{itemize}

\item \textbf{Language Identification}: Currently, MinerU identifies and processes only Chinese and English documents. The language type needs to be specified as a parameter when performing OCR, and the quality of processing for other languages is not guaranteed.

\item \textbf{Content Garbled Detection}: Some text-based PDFs contain text that appears garbled when copied. Such PDFs need to be identified in advance so that OCR can be used for text recognition in the next step.

\item \textbf{Scanned PDF Identification}: For text-based PDFs (as opposed to scanned PDFs), MinerU directly uses PyMuPDF for text extraction. However, for scanned PDFs, OCR needs to be enabled. Scanned PDFs are identified based on characteristics such as a larger image area compared to the text area, sometimes covering the entire PDF page, and an average text length per page close to zero.

\item \textbf{Page Metadata Extraction}: Extracts document metadata such as total page count, page dimensions (width and height), and other relevant attributes.

\end{itemize}

\subsection{Document Content Parsing}

In the document parsing stage, MinerU uses the PDF-Extract-Kit model library to detect different types of regions  and recognize the corresponding region contents (OCR, formula recognition, table recognition, etc.). PDF-Extract-Kit is an algorithm library for PDF parsing, containing various state-of-the-art (SOTA) open-source PDF document parsing algorithms. Unlike other open-source algorithm libraries, PDF-Extract-Kit aims to build a model library that ensures accuracy and speed when dealing with diverse data in real-world scenarios. When the SOTA open-source algorithms in a specific field fail to meet practical needs, PDF-Extract-Kit employs data engineering to construct high-quality, diverse datasets for further model fine-tuning, thereby significantly enhancing the model's robustness to varied data. The current version of MinerU\footnote{Current version: v0.8.1} utilizes five models: layout detection, formula detection, table recognition, formula recognition and OCR.

\begin{figure*}[ht]
\centering
\includegraphics[width=0.98\linewidth]{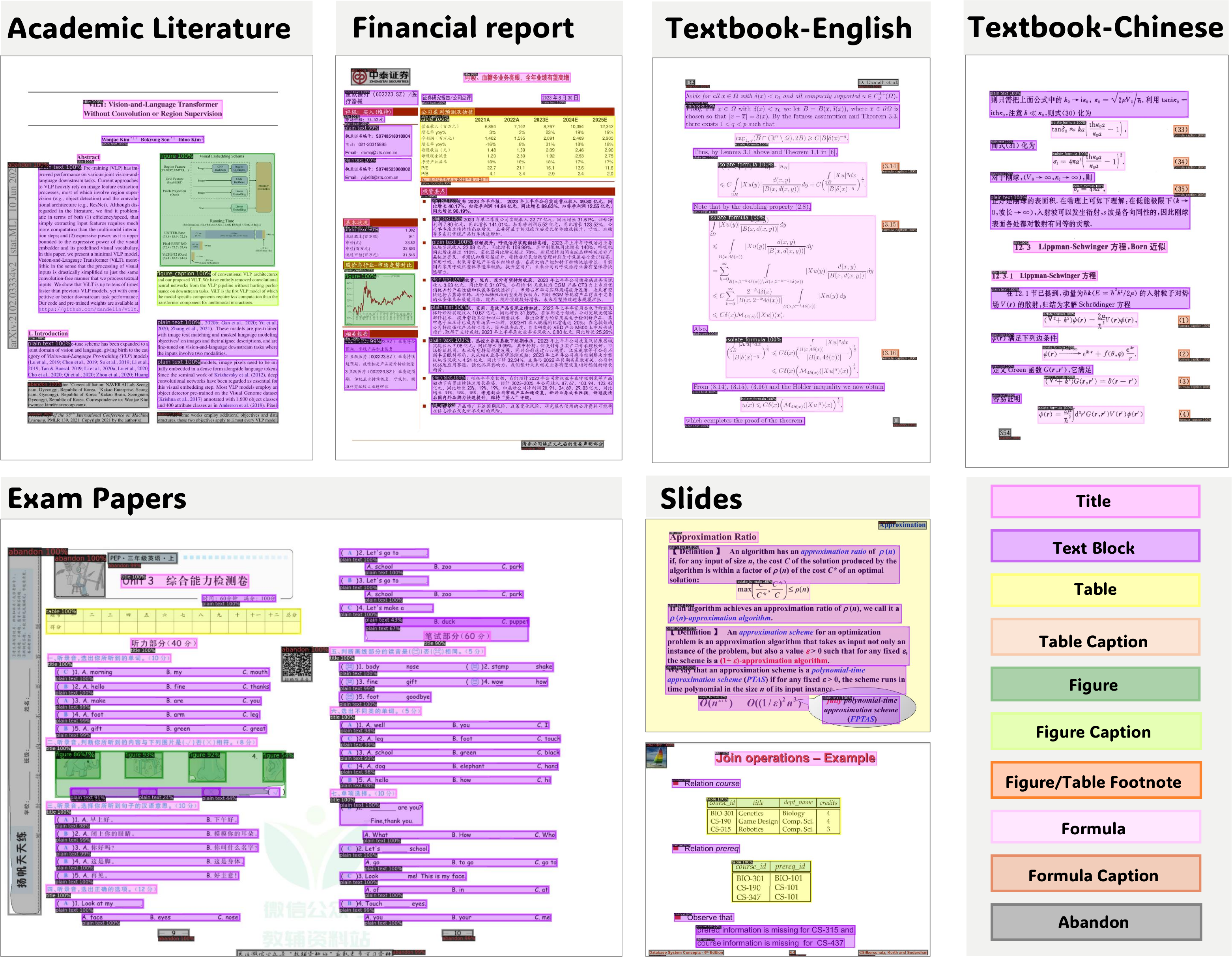}
    \caption{High-quality layout detection results on diverse documents.}
\label{fig:fig2}
\vspace{-10pt}
\end{figure*}

\subsubsection{Layout Analysis}

Layout analysis is the crucial first step in document parsing, aiming to distinguish different types of elements and their corresponding regions on a page. Existing layout detection algorithms ~\cite{huang2022layoutlmv3, yao2023docxchain} perform well on paper-type documents but struggle with diverse documents such as textbooks and exam papers. Therefore, PDF-Extract-Kit constructs a diverse layout detection training set and trains high-quality models for document region extraction.

The data engineering-based model training approach is as follows:

\begin{itemize}
    \item \textbf{Diverse Data Selection}: Collects diverse PDF documents, clusters them based on visual features, and samples data from different cluster centers to obtain an initial diverse document dataset. The categories include scientific papers, general books, textbooks, exam papers, magazines, PPTs, research reports, etc.
    
    \item \textbf{Data Annotation}: Categorizes the layout annotation types involved in the document components, including titles, body paragraphs, images, image captions, tables, table captions, image table notes, inline formulas, formula labels, and discard types (such as headers, footers, page numbers, and page notes). Detailed annotation standards are established for each type, and approximately 21K data points are annotated as the training set.
    
    \item \textbf{Model Training}: Fine-tunes the model for the Layout Detection task based on the layout detection models~\cite{huang2022layoutlmv3,wang2024yolov10}. The number of classes parameter is modified to align with our categorized layout types.

    \item \textbf{Iterative Data Selection and Model Training}: During model iteration, partitions a portion of the data as a validation set and uses its results to guide the focus of subsequent data iterations. If a specific category from a particular source of PDF documents scores low, the sampling weight for PDF pages containing that specific category from that source is increased in the next iteration, thereby more efficiently iterating the data and model.
    
\end{itemize}

The model trained on diverse datasets performs significantly better on varied documents. As shown in \Cref{fig:fig2}, the layout detection model trained on diverse layout detection data performs well on documents such as textbooks, far exceeding the performance of open-source SOTA models.

\subsubsection{Formula Detection}

Layout analysis can accurately locate most elements in a document, but formula types, especially inline formulas, can be visually indistinguishable from text, such as "$100cm^2$" and "\((\alpha_1, \alpha_2, \ldots, \alpha_n)\)". If formulas are not detected in advance, subsequent text extraction using OCR or Python libraries may result in garbled text, affecting the overall accuracy of the document, which is crucial for scientific documents. Therefore, we trained a dedicated formula detection model.

For the formula detection dataset annotation, we defined three categories: inline formulas, displayed formulas, and an ignore class. The ignore class mainly refers to areas that are difficult to determine as formulas, such as "50\%", "NaCl", and "1-2 days". Ultimately, we annotated 24,157 inline formulas and 1,829 displayed formulas on 2,890 pages from Chinese and English papers, textbooks, books, and financial reports for training.

After obtaining a diverse formula detection dataset, PDF-Extract-Kit trains a YOLO-based model, which performs well in terms of speed and accuracy on various documents.

\subsubsection{Formula Recognition}

Varied documents contain various types of formulas, such as short printed inline formulas and complex displayed formulas. Some documents are scanned, leading to noisy formula content and even the presence of handwritten formulas. Therefore, MinerU employs the self-developed UniMERNet~\cite{wang2024unimernet} model for formula recognition. The UniMERNet model is trained on the large-scale diverse formula recognition dataset UniMER-1M. Thanks to the optimization of the model structure, it achieves good performance on various types of formulas (SPE, CPE, SCE, HWE) in real-world scenarios, comparable to commercial software MathPix~\cite{mathpix2024,wang2024cdm}.

\subsubsection{Table Recognition}
Tables serve as an effective way to present structured data across various contexts, including scientific publications, financial reports, invoices, web pages, and beyond. Extracting tabular data from visual table images, known as the table recognition task, is challenging primarily because tables often contain complex column and row headers, as well as spanning cell operations. By leveraging MinerU, users can perform Table-to-LaTex or Table-to-HTML tasks to extract structured data from tables. MinerU employs TableMaster~\cite{ye2021pingan} and StructEqTable\footnote{\url{https://github.com/UniModal4Reasoning/StructEqTable-Deploy}} for performing the table recognition task. TableMaster is trained using PubTabNet dataset (v2.0.0)~\cite{zhong2020image} while StructEqTable is trained using data from DocGenome benchmark~\cite{xia2024docgenome}. TableMaster divides the table recognition task into four sub-tasks including table structure recognition, text line detection, text line recognition, and box assignment, while StructEqTable performs the table recognition task in an end-to-end manner, demonstrating stronger recognition performance and delivering good results even with complex tables.

\subsubsection{OCR}

\begin{figure*}[ht]
\centering
    \includegraphics[width=1.0\linewidth]{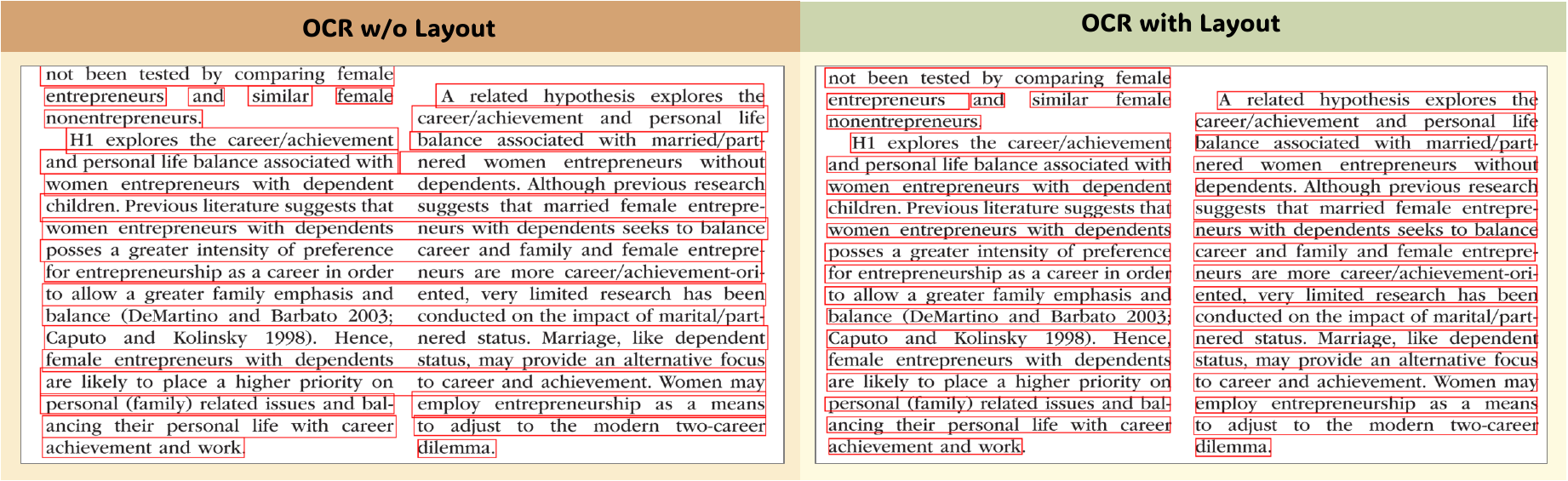}
    \caption{OCR results comparison on a multi-column document. The left image shows incorrect text order without layout detection, while the right image preserves the correct order with layout detection.}
\label{fig:fig3}
\vspace{-5pt}
\end{figure*}

\begin{figure*}[ht]
\centering
    \includegraphics[width=1.0\linewidth]{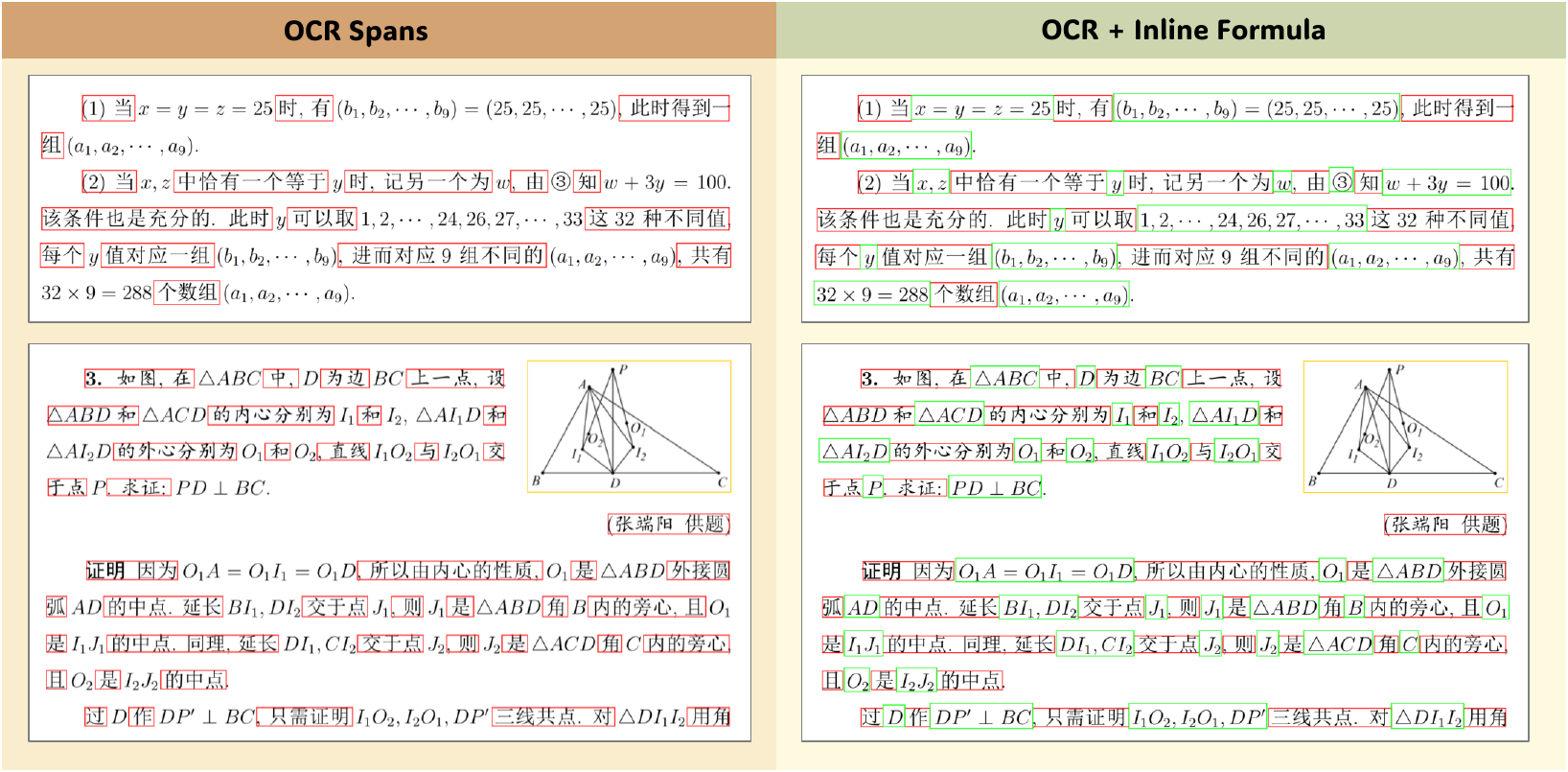}
    \caption{OCR results for text blocks with inline formulas. The left image shows OCR results with formulas masked, and the right image shows the final results with formulas reintegrated.}
\label{fig:fig4}
\vspace{-10pt}
\end{figure*}

After excluding special regions (tables, formulas, images, etc.) in the document, we can directly apply OCR to recognize text regions. MinerU uses Paddle-OCR\footnote{\url{https://github.com/PaddlePaddle/PaddleOCR}} integrated into PDF-Extract-Kit for text recognition. However, as shown in \Cref{fig:fig3}, direct OCR on the entire page can sometimes result in text from different columns being recognized as a single column, which leads to incorrect text order. Therefore, we perform OCR based on the text regions (titles, text paragraphs) detected by the layout analysis to avoid disrupting the reading order.

As shown in \Cref{fig:fig4}, When performing OCR on text blocks with inline formulas, we first mask the formulas using the coordinates provided by the formula detection model, then perform OCR, and finally reinsert the formulas back into the OCR results.

\subsection{Document Content Post-Processing}

\begin{figure*}[ht]
    \includegraphics[width=1.0\linewidth]{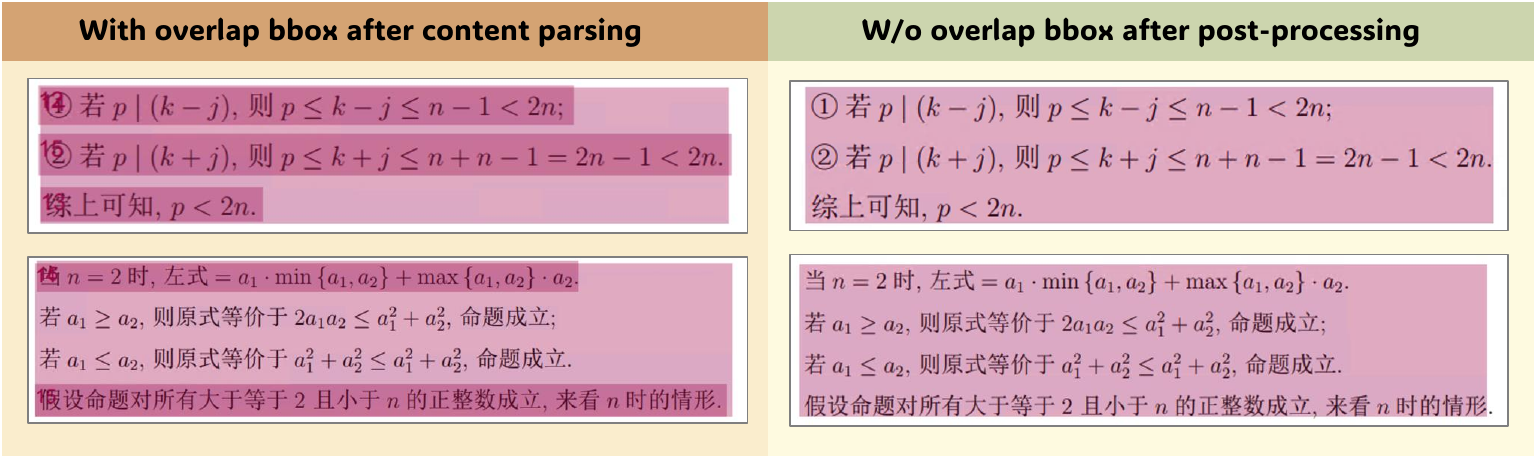}
    \caption{Bounding Boxes before and after resolving overlaps. The left image shows overlapping BBoxes, and the right image shows the results after removing overlaps.}
\label{fig:fig5}
\end{figure*}

The post-processing stage primarily addresses the issue of content ordering. Due to potential overlaps among text, images, tables, and formula boxes output by the model, as well as frequent overlaps among text lines obtained through OCR or API, sorting the text and elements poses a significant challenge. This stage focuses on handling the relationships between Bounding Boxes (BBox). \Cref{fig:fig5} shows a visualization of the results before and after resolving overlapping bounding boxes.

The solutions to the BBox relationships include the following aspects:

\textbf{Containment Relationships}. Remove formulas and text blocks contained within image and table regions, as well as boxes contained within formula boxes.

\textbf{Partial Overlap Relationships}. Partially overlapping text boxes are shrunk vertically and horizontally to avoid mutual coverage, ensuring that the final position and content remain unaffected, which facilitates subsequent sorting operations. For partial overlaps between text and tables/images, the integrity of the text is ensured by temporarily ignoring the images and tables.

After addressing the nested and partially overlapping BBoxes, MinerU developed a segmentation algorithm based on the human reading order, "top to bottom, left to right." This algorithm divides the entire page into several regions, each containing multiple BBoxes, while ensuring that each region contains at most one column. This ensures that the text is read line by line from top to bottom, adhering to the natural human reading sequence. The segmented groups are then sorted according to their positional relationships, determining the reading order of each element within the PDF.

\begin{figure*}[t]
    \includegraphics[width=0.98\linewidth]{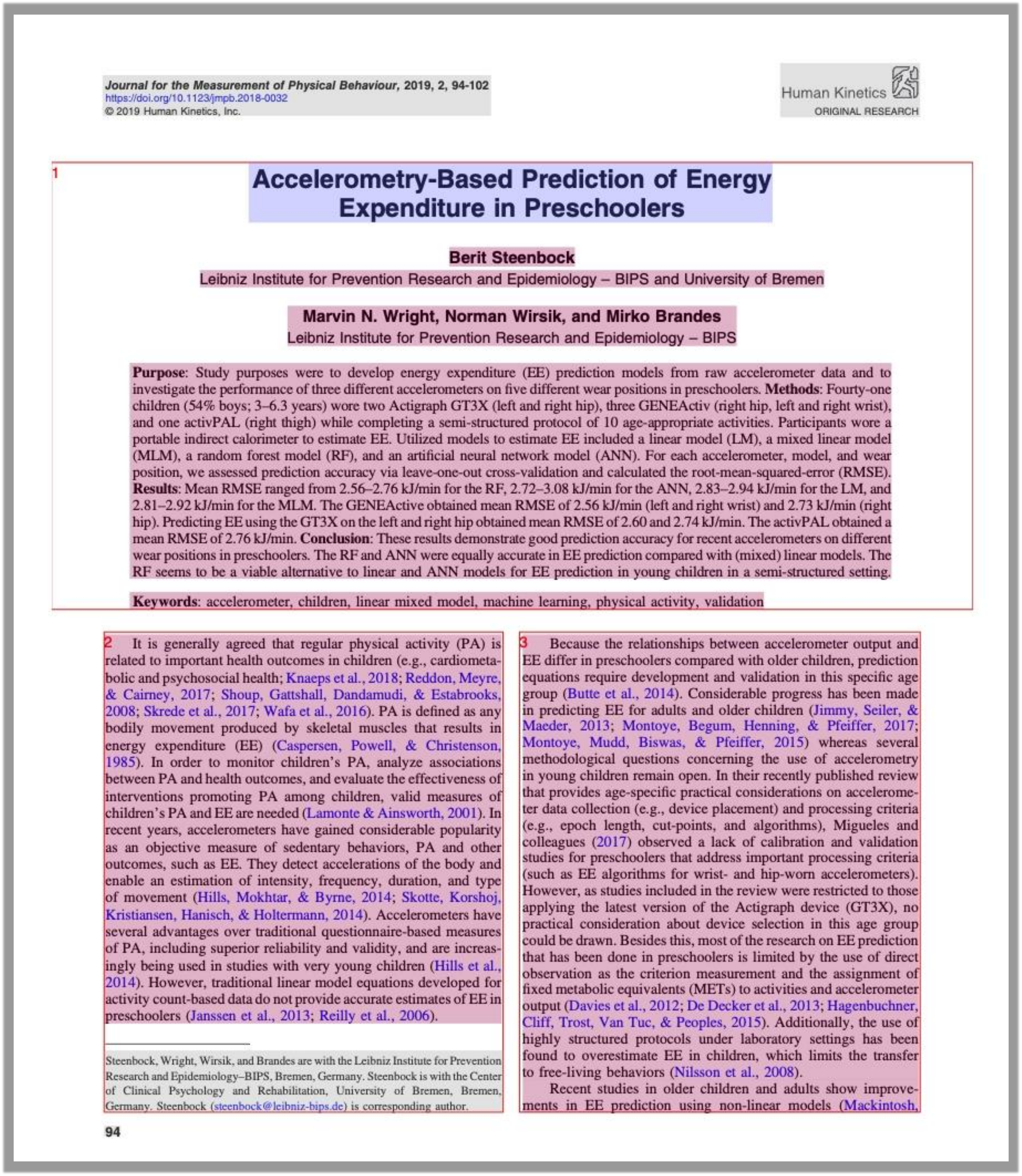}
    \caption{Visualization of the region sorting results.}
\label{fig:fig6}
\vspace{-15pt}
\end{figure*}

\subsection{Format Conversion}
To accommodate varying user requirements for output formats, MinerU stores the processed PDF data in an intermediate structure. The intermediate structure is a large JSON file,with the most important fields listed in \Cref{tab:tab1}.

\begin{table*}[h]
\centering
\resizebox{0.98\linewidth}{!}{
\setlength{\tabcolsep}{6pt}
{
\begin{tabular}{l|p{12cm}}
\toprule[0.9pt]
\textbf{Field Name} & \textbf{Function} \\
\midrule
pdf\_info & This field contains multiple subfields. The most important one is para\_blocks, an ordered array where each element represents a segment of content on the PDF, which can be images, image captions, text, titles, tables, etc. Concatenating the content of this array in order reconstructs the content of the PDF (excluding headers, footers, page numbers, etc.). \\
\midrule
\_parse\_type & Takes values of txt or ocr. If it is txt, it means the text is directly extracted from the PDF via API. If it is ocr, it means the text is obtained through an OCR engine. \\
\midrule
\_version\_name & The software version, which can be used to track errors in data processing. \\
\bottomrule[0.9pt]
\end{tabular}
}}
\caption{Important Fields in the Intermediate Structure}
\label{tab:tab1}
\end{table*}

MinerU's command line supports output in Markdown and a custom JSON format, both converted from the aforementioned intermediate structure. During the format conversion process, images, tables, and other elements can be cropped as needed. For detailed format descriptions, refer to the documentation\footnote{\url{https://github.com/opendatalab/MinerU/blob/master/docs/output_file_en_us.md}}.

\begin{table*}[ht]
\centering
\resizebox{\linewidth}{!}{
\setlength{\tabcolsep}{6pt}
{
\begin{tabular}{l|p{12cm}}
\toprule[0.9pt]
\textbf{Category} & \textbf{Description} \\
\midrule
Research Report & Financial reports from the internet, featuring large tables, complex merged tables, horizontal tables mixed with text, single and double columns, and complex layouts. \\
\midrule
Standard Textbook & Textbooks from the internet, characterized by single-column layout, black-and-white color, nested complex formulas, and large matrices. \\
\midrule
Special Image-Text Textbook & Textbooks from the internet with special image-text content, covering subjects like English, Mathematics, and Chinese (including Pinyin). \\
\midrule
Academic Paper & Documents from arXiv and SCIHUB, featuring complex layouts with single and double columns, figures, tables, and formulas. \\
\midrule
Picture Album & Picture albums from the internet, characterized by pages with large images. \\
\midrule
PowerPoint Slides & PDF files converted from internet PowerPoint slides, featuring background colors and covering subjects like Biology, Chinese, English, and Physics. \\
\midrule
Standard Exam Paper & Exam papers from the internet, characterized by exam layout, black-and-white background, and covering subjects like Computer Science, Mathematics, and Chinese, including primary, middle, high school, and industry question banks. \\
\midrule
Special Image-Text Exam Paper & Exam papers from the internet with special image-text content, covering subjects like English, Mathematics, and Chinese (including Pinyin). \\
\midrule
Historical Document & Documents from the internet, characterized by vertical layout, right-to-left reading order, and traditional Chinese fonts. \\
\midrule
Notes & Notes from the internet, featuring handwritten content, including notes from three middle school students. \\
\midrule
Standard Book & Books from the internet, characterized by single-column layout and black-and-white background. \\
\bottomrule[0.9pt]
\end{tabular}
}}
\caption{Categories of Documents and Their Descriptions}
\label{tab:document_categories}
\end{table*}

\section{MinerU Quality Assessment}

To assess the quality of content extracted by MinerU from PDFs, we explore two dimensions. First, we conduct a standalone evaluation of the core modules responsible for document content parsing to ensure the accuracy of model inference results. The quality of model results is crucial for the final content quality, as evidenced by the overall process. At this stage, we specifically evaluate three modules: layout detection, formula detection, and formula recognition. We construct a diverse evaluation dataset and compare the performance of the core algorithm components of MinerU's PDF-Extract-Kit with other state-of-the-art (SOTA) open-source models. Additionally, we perform manual quality checks to assess MinerU's performance on diverse document types.

\subsection{Construction of a Diverse Evaluation Dataset}

To assess the quality of document content extraction in real-world scenarios, we initially constructed a diverse evaluation dataset for model assessment and visual analysis of extracted content. As shown in \Cref{tab:document_categories}, the diverse dataset includes 11 types of documents, from which we further construct evaluation datasets for layout detection and formula detection.

\begin{table*}[ht]
\centering
\resizebox{0.9\linewidth}{!}{
\setlength{\tabcolsep}{6pt}
{
\begin{tabular}{lcccccc}
\toprule[0.9pt]
\multirow{2}{*}{\textbf{Model}} & \multicolumn{3}{c}{\textbf{Academic Papers Val}} & \multicolumn{3}{c}{\textbf{Textbook Val}} \\
\cmidrule(lr){2-4} \cmidrule(lr){5-7}
 & \textbf{mAP} & \textbf{AP50} & \textbf{AR50} & \textbf{mAP} & \textbf{AP50} & \textbf{AR50} \\
\midrule
DocXchain & 52.8 & 69.5 & 77.3 & 34.9 & 50.1 & 63.5 \\
Surya & 24.2 & 39.4 & 66.1 & 13.9 & 23.3 & 49.9 \\
360LayoutAnalysis-Paper & 37.7 & 53.6 & 59.8 & 20.7 & 31.3 & 43.6 \\
360LayoutAnalysis-Report & 35.1 & 46.9 & 55.9 & 25.4 & 33.7 & 45.1 \\
LayoutLMv3-Finetined (Ours) & \textbf{77.6} & \textbf{93.3} & \textbf{95.5} & \textbf{67.9} & \textbf{82.7} & \textbf{87.9} \\
\bottomrule[0.9pt]
\end{tabular}
}}
\caption{Performance of different models on layout detection}
\label{tab:layout_detection_performance}
\end{table*}

\subsection{Evaluation of Core Algorithm Modules}

\subsubsection{Layout Detection}

We compared MinerU's layout detection model with existing open-source models, including DocXchain~\cite{yao2023docxchain}, Surya\footnote{\url{https://github.com/VikParuchuri/surya}}, and two models from 360LayoutAnalysis\footnote{\url{https://github.com/360AILAB-NLP/360LayoutAnalysis}}. \Cref{tab:layout_detection_performance} shows the performance of each model on academic papers and textbook validation sets. The LayoutLMv3-SFT model, as shown in the table, was fine-tuned on our internally constructed layout detection dataset based on the LayoutLMv3-base-chinese pretrained model. The initial evaluation dataset for layout detection includes validation sets from academic papers and textbooks.

\begin{table*}[ht]
\centering
\resizebox{0.8\linewidth}{!}{
\setlength{\tabcolsep}{6pt}
{
\begin{tabular}{lcccc}
\toprule[0.9pt]
\multirow{2}{*}{\textbf{Model}} & \multicolumn{2}{c}{\textbf{Academic Papers Val}} & \multicolumn{2}{c}{\textbf{Multi-source Val}} \\
\cmidrule(lr){2-3} \cmidrule(lr){4-5}
 & \textbf{AP50} & \textbf{AR50} & \textbf{AP50} & \textbf{AR50} \\
\midrule
Pix2Text-MFD & 60.1 & 64.6 & 58.9 & 62.8 \\
YOLOv8-Finetined (Ours) & \textbf{87.7} & \textbf{89.9} & \textbf{82.4} & \textbf{87.3} \\
\bottomrule[0.9pt]
\end{tabular}
}}
\caption{Performance of different models on formula detection}
\label{tab:formula_detection_performance}
\end{table*}

\subsubsection{Formula Detection}

We compare MinerU's formula detection model with the open-source formula detection model Pix2Text-MFD. Additionally, YOLO-Finetuned is a model we trained based on YOLOv8 using a diverse formula detection training set.

The formula detection evaluation dataset comprises pages from academic papers and various sources for formula detection. The results, as shown in \Cref{tab:formula_detection_performance}, demonstrate that the detection model fine-tuned on diverse data significantly outperforms previous open-source models on both papers and various other document types.


\begin{table*}[ht]
\centering
\resizebox{0.75\linewidth}{!}{
\setlength{\tabcolsep}{6pt}
{
\begin{tabular}{lcccc}
\toprule[0.9pt]
 \textbf{Model} & \textcolor{gray}{\textbf{ExpRate}} & \textbf{ExpRate@CDM} & \textcolor{gray}{\textbf{BLEU}} & \textbf{CDM} \\
 \midrule
Pix2tex & \textcolor{gray}{0.1237} & 0.291 & \textcolor{gray}{0.4080} & 0.636 \\
Texify & \textcolor{gray}{0.2288} & 0.495 & \textcolor{gray}{0.5890} & 0.755 \\      
Mathpix & \textcolor{gray}{0.2610} & 0.5 & \textcolor{gray}{0.8067} & 0.951 \\
UniMERNet & \textcolor{gray}{0.4799} & 0.811 & \textcolor{gray}{0.8425} & 0.968 \\
\bottomrule[0.9pt]
\end{tabular}
}}
\caption{Evaluation results of different models on the UniMER-Test dataset. Results are adapted from the CDM paper~\cite{wang2024cdm}. The ExpRate and BLEU metrics are shown in gray as they are considered less reliable. The CDM metric is unaffected by the diversity of formula representations and is therefore a more reasonable metric for comparing the formula recognition performance of different models.}
\label{tab:tab5}
\end{table*}

\subsubsection{Formula Recognition}

PDFs contain various types of formulas, and to achieve robust formula recognition results on diverse formulas, we use UniMERNet as our formula recognition model. Given that the same formula may have various expressions, we utilize CDM~\cite{wang2024cdm} for evaluating formula recognition performance. As shown in \Cref{tab:tab5}, UniMERNet's formula recognition capability far surpasses that of other open-source models and is comparable to commercial software like Mathpix.

Based on the above evaluations, we can conclude that the models used by MinerU, trained specifically on diverse document sources, significantly outperform other open-source models designed for single document types, ensuring the accuracy of parsing results.

\begin{figure*}[t]
\centering
    \includegraphics[width=1.0\linewidth]{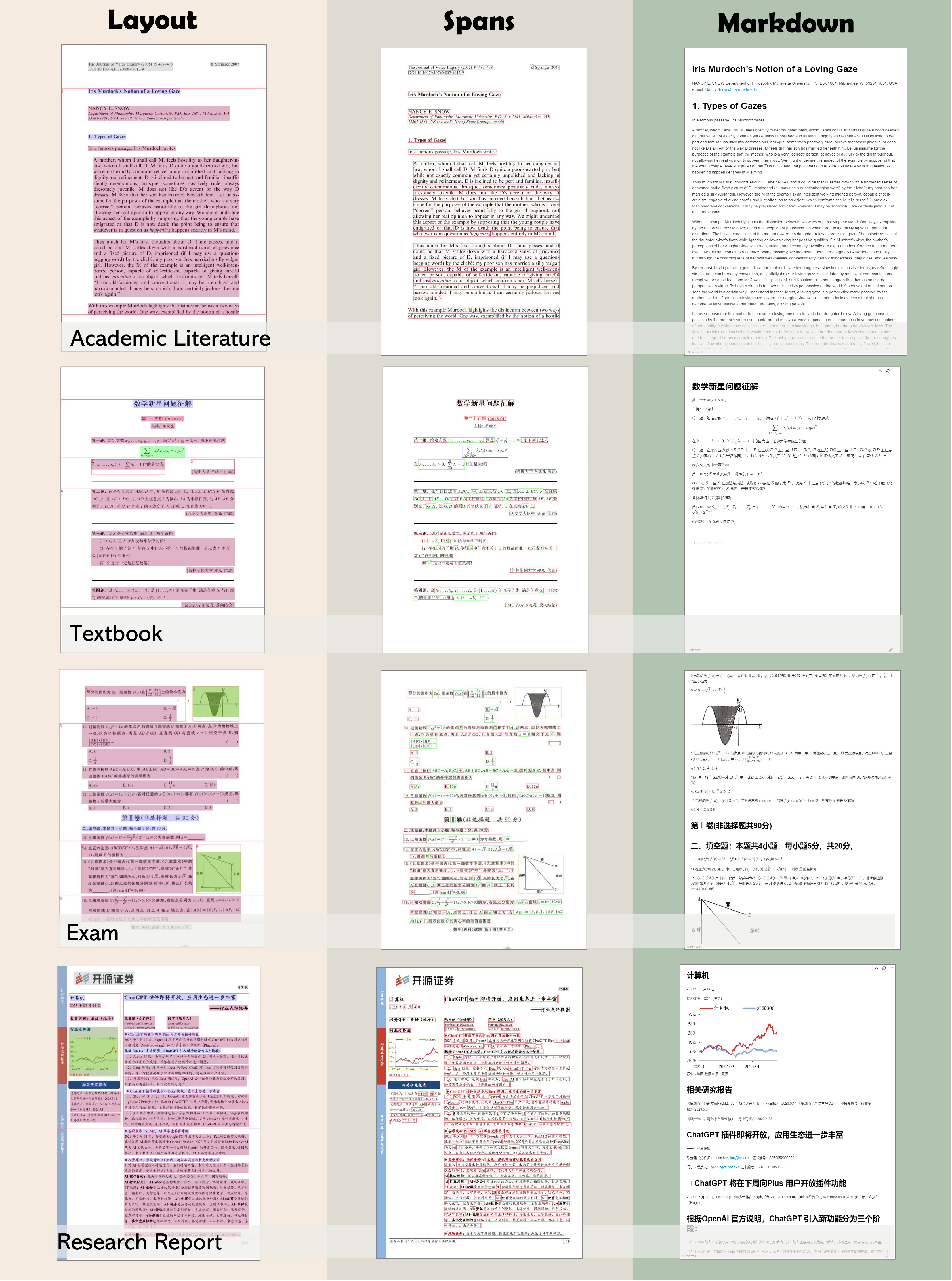}
\caption{Visualization of MinerU's extraction process on various document types. From left to right: layout detection results, span results, and final Markdown results.}

\label{fig:fig7}
\end{figure*}

\subsection{End-to-End Results Visualization and Analysis}

To assess the quality of MinerU's final extraction results, in addition to ensuring the quality of the model extraction results mentioned above, we also perform post-processing on the extracted results, such as removing noise content and stitching model outputs. MinerU's post-processing operations ensure the readability and accuracy of the final results. As shown in \Cref{fig:fig7}, MinerU achieves excellent extraction results on diverse documents.

From the visualization results, it is evident that the layout detection results accurately locate different regions. The spans \footnote{In document extraction tasks, a span is often used to mark and process specific text segments} show that the formula detection and OCR detection results are satisfactory, ultimately stitching together into high-quality Markdown results.

\section{Conclusion and Future Work}

In this work, we introduce MinerU, a one-stop PDF document extraction tool. Thanks to high-quality model inference results and meticulous pre-processing and post-processing operations, MinerU ensures high-quality extraction results even when dealing with diverse document types. Although MinerU has demonstrated significant advantages, there is still ample room for improvement. Moving forward, we continuously upgrade MinerU in the following areas:

\begin{itemize}
    \item \textbf{Enhancement of Core Components}. We will iteratively update the existing models in the PDF-Extract-Kit to further improve the extraction quality for diverse documents. Additionally, we will introduce new models, such as table recognition and reading order, to enhance MinerU's overall capabilities.
    
    \item \textbf{Improvement of Usability and Inference Speed}. We will further optimize MinerU's processing pipeline to accelerate document extraction speed and enhance usability. Moreover, we will deploy more efficient online inference services to meet users' real-time needs.
    
    \item  \textbf{Systematic Benchmark Construction}. We will establish a systematic evaluation benchmark for diverse documents to clearly compare the results of MinerU with those of state-of-the-art open-source methods, aiding community users in selecting the most suitable models for their needs.

\end{itemize}

\clearpage

\bibliographystyle{plain}
\bibliography{mineru}

\end{document}